%% file: main.tex
\documentclass{article}
\usepackage{ijcai20}

\usepackage{placeins}

\usepackage{times}

\usepackage{soul}
\usepackage{url}
\usepackage[hidelinks]{hyperref}
\usepackage[utf8]{inputenc}
\usepackage[small]{caption}
\usepackage{graphicx}
\usepackage{amsmath}
\usepackage{booktabs}
\input{math_commands.tex}

\usepackage{amssymb}
\usepackage[normalem]{ulem}
\usepackage{graphicx}
\usepackage{todonotes}
\definecolor{mygray}{gray}{0.8}
\definecolor{dgreen}{rgb}{0,0.6,0}
\hypersetup{colorlinks,citecolor=dgreen,urlcolor=blue}
\usepackage{algorithm}
\usepackage{algorithmic}
\usepackage{diagbox}
\usepackage{multirow}

\hypersetup{
    colorlinks = false,
    linkbordercolor = {white}
}

\usepackage{mathtools}

\usepackage{makecell}

\newcommand{\Prange}[2]{{P_{#1}^{#2}} }

\newcommand{\nostro}{MARTHE}  

\title{\nostro{}: Scheduling the Learning Rate Via Online Hypergradients}

\author{
Michele Donini$^{1}$\footnote{Contact Author, Equal Contribution.}\and
Luca Franceschi$^{2,3}$\footnote{Contact Author, Equal Contribution.}\and
 Orchid Majumder$^{1}$
\and \\
 Massimiliano Pontil$^{2,3}$
 \And
 Paolo Frasconi$^4$\\
\affiliations
$^1$Amazon \\
 $^2$University College London, London, UK \\
 $^3$Istituto Italiano di Tecnologia, Genova, Italy \\
 $^4$Università di Firenze, Firenze, Italy \\
 \vspace{1mm}
\emails
donini@amazon.com; 
luca.franceschi@iit.it
}

\begin{document}

\maketitle

\begin{abstract}
  We study the problem of fitting task-specific learning rate
  schedules from the perspective of hyperparameter optimization,
  aiming at good generalization. 
  We describe the structure of the gradient of a validation error w.r.t. the learning rate schedule -- the hypergradient. Based on this, we introduce \nostro{}, a novel online algorithm guided by cheap approximations of the hypergradient that uses past information from the optimization trajectory to simulate future behaviour. It interpolates between two recent
  techniques, RTHO \cite{franceschi2017forward} and HD
  \cite{baydin_2017_online}, and is able to produce learning rate schedules that are more stable leading to  models that generalize better.
\end{abstract}

\section{Introduction}
Learning rate (LR) adaptation for first-order optimization methods is
one of the most widely studied aspects in optimization for learning
methods, in particular neural networks.
Recent research in this area has focused on developing complex schedules that depend
on a small number of hyperparameters~\cite{loshchilov2017sgdr,orabona2016coin} or proposed methods to optimize LR schedules w.r.t. the training
objective~\cite{schaul2013no,baydin_2017_online,wu2018understanding}. While
quick optimization is desirable, the true goal of supervised learning
is to minimize the generalization error, which is commonly estimated
by holding out part of the available data for
validation. Hyperparameter optimization (HPO), a related but distinct
branch of the literature, specifically focuses on this aspect, with
less emphasis on the goal of rapid convergence on a single task.
Research in this direction is vast and includes model-based, model-free, and gradient-based approaches
(see~\cite{hutter_automl_2019} for
an overview).
Additionally, works in the area of learning to
optimize~\cite{andrychowicz2016learning,wichrowska2017learnedICML}
have focused on the problem of tuning parameterized optimizers on
whole classes of learning problems but require prior expensive
optimization and are not designed to speed up training on a single task.

The goal of this paper is to automatically compute \textit{in an
  online fashion} a learning rate schedule for stochastic optimization
methods (such as SGD) only on the basis of the given learning task,
aiming at producing models with associated small validation error.
We study the problem of finding a LR schedule under the framework of
gradient-based hyperparameter
optimization~\cite{franceschi2017forward}: we consider an optimal
schedule $\eta^* = (\eta^*_0,\dots,\eta^*_{T-1})\in\mathbb{R}^T_+$ as a
solution to the following constrained optimization problem
\begin{align}
    \label{eq:main}
    \min \{f_T(\eta) =E(w_T(\eta)) : \eta \in \mathbb{R}_+^T\} \\ 
    \text{s.t.} \quad w_0 = \bar{w}, \quad w_{t+1}(\eta)=\Phi_t(w_{t}(\eta),\eta_t) \nonumber
\end{align}
for $t=\{0, \dots, T-1\}=[T]$, where $E:\mathbb{R}^d\to\mathbb{R}_+$ is an
objective function, $\Phi_t:\mathbb{R}^d\times \mathbb{R}_+\to\mathbb{R}^d$ is a
(possibly stochastic) weight update dynamics, $\bar{w}\in\mathbb{R}^d$
represents the initial model weights (parameters) and finally $w_t$ are the
weights after $t$ iterations. 
We can think of $E$ as either the training or the validation loss of
the model, while the dynamics $\Phi$ describe the update rule (such as
SGD, SGD-Momentum, Adam etc.). For example in the case of SGD,
$\Phi_t(w_t,\eta_t)=w_t-\eta_t\nabla L_t(w_t)$, with $L_t(w_t)$ the training loss
on the $t$-th minibatch. The \emph{horizon} $T$ should be large enough so that
the training error can be effectively minimized to avoid underfitting.
A too large value of $T$ does not necessarily harm since $\eta_k=0$
for $k>\bar{T}$ is still a feasible solution, implementing early stopping in 
this setting.
There is vast empirical evidence that  the LR is among the most critical hyperparameters affecting the performances of learnt statistical models. Beside convergence arguments from the stochastic optimization literature, for all but the simplest problems, non-constant schedules yield generally better results \cite{bengio2012practical}.

Problem (\ref{eq:main}) can be in principle solved by any HPO technique.  
However, most HPO techniques, including those based on
hypergradients~\cite{maclaurin_gradient-based_2015} or on a bilevel programming
formulation~\cite{franceschi2018bilevel,mackay2019self} would not be suitable
for the present purpose since they require multiple evaluations of $f$ (which,
in turn, require executions of the weight optimization routine). This clearly defeats 
one of the main goals of determining LR schedules, i.e. speed. In fact, several
other
researchers~\cite{almeida1999parameter,schraudolph1999local,schaul2013no,%
franceschi2017forward,baydin_2017_online,wu2018understanding} have investigated
related solutions for deriving greedy update rules for the learning rate.  A
common characteristic of methods in this family is that the LR update rule does
not take into account information from the future.  
At a high level, we argue that any method should attempt to
produce updates that approximate the true and computationally unaffordable
hypergradient of the \textit{final} objective with respect to the current
learning rate. In relation to this,~\cite{wu2018understanding} discuss the
bias deriving from greedy or short-horizon optimization.
In practice, different methods resort to different approximations or explicitly
consider greedily minimizing the performance after a single parameter update
\cite{almeida1999parameter,schaul2013no,baydin_2017_online}.
The type of approximation and the type of objective (i.e. the training or the
validation loss) are in principle separate issues although comparative
experiments with both objectives and the same approximation are never reported
in the literature. The validation loss is only used in experiments
reported by Franceschi \emph{et al.}~\shortcite{franceschi2017forward}.

 In this work, we make a step forward in
 understanding the behavior of online gradient-based hyperparameter optimization
 techniques by (i) analyzing in Section~\ref{sec:structure} the structure of
 the true hypergradient that could be used to solve Problem~(\ref{eq:main}) if
 wall-clock time was not a concern,  and (ii) by studying in
 Section~\ref{sec:HD_RTHO_modes} some failure modes of previously proposed
 methods along with a detailed discussion of the type of approximations that these
 methods exploit. In Section~\ref{sec:proposal}, based on these considerations,
 we develop a new hypergradient-based algorithm 
which reliably 
produces competitive learning rate schedules aimed at lowering the final
validation error. The algorithm, which we call \nostro{} (Moving Average Real-Time Hyperparameter Estimation),  has a moderate computational cost and can be
interpreted as a generalization of the algorithms described by Baydin \emph{et al.}~\shortcite{baydin_2017_online}
and Franceschi \emph{et al.}~\shortcite{franceschi2017forward}. In Section~\ref{sec:true-vs-approx}, we empirically
compare the quality of different hypergradient approximations in a small scale task where true hypergradient can be exactly computed. In Section~\ref{sec:exp}, we
present a set of real world experiments showing the validity of our approach. We finally
discuss potential future applications and research directions in
Section~\ref{sec:conclusions}. 

\section{Structure of the Hypergradient} \label{sec:structure}
We study the optimization problem~(\ref{eq:main})  under the perspective 
of gradient-based hyperparameter optimization, where the learning rate
schedule $\eta=(\eta_0,\dots,\eta_{T-1})$
is treated as a vector of hyperparameters and $T$ is a fixed horizon. 
Since the learning rates are positive real-valued variables, assuming both $E$ and $\Phi$ are smooth functions, we can compute the gradient of $f\in\mathbb{R}^{T}$, which is given by
\begin{equation} \label{eq:hypergrad:opt}
    \nabla f_T(\eta) = \dot{w}_T^\intercal \nabla E(w_T),  \quad \dot{w}_T = \frac{\mathrm{d} w_T}{\mathrm{d} \eta} \in \mathbb{R}^{d\times T },
    \end{equation}
where ``$\intercal$'' means transpose. The total derivative 
$\dot{w}_T$ can be computed iteratively with forward-mode algorithmic differentiation \cite{griewank2008evaluating,franceschi2017forward} as
\begin{align} \label{eq:wdot:upd1}
    &\dot{w}_0 = 0, \quad \dot{w}_{t+1} = A_t\dot{w}_{t} + B_t,   \\
    \text{ with } \qquad & A_t = \frac{\partial \Phi_t(w_{t}, \eta_t)}{\partial w_t},
    \quad 
    B_t = \frac{\partial \Phi_t(w_{t}, \eta_t)}{\partial \eta}.
\end{align}
The Jacobian matrices $A_t$ and $B_t$ depend on $w_{t}$ and $\eta_t$, but we will leave these dependencies implicit to ease our notation. In the case of SGD\footnote{Throughout we use SGD to simplify the discussion, however, similar arguments hold for any smooth optimization dynamics such as those including momentum terms.},
    $A_t = I - \eta_t H_{t}(w_{t})$ and $[B_t]_j = -\delta_{tj} \nabla L_t(w_{t})^\intercal$, where subscripts denote columns (starting from 0),
$\delta_{tj} = 1$ if $t=j$ and $0$ otherwise and $H_t$ is the Hessian of the training error on the $t-$th mini-batch.

Given the high dimensionality of $\eta$, reverse-mode
differentiation would result in a more efficient (running-time)
implementation. We use here forward-mode both because it is easier to interpret 
and because it is closely related to the computational scheme behind \nostro, as we will show in Section \ref{sec:proposal}.
We note that stochastic approximations of Eq. (\ref{eq:hypergrad:opt}) may be obtained with randomized telescoping sums \cite{beaston2019efficient} or hyper-networks based stochastic approximations \cite{mackay2019self}.

Eq. (\ref{eq:wdot:upd1}) describes the so-called tangent system \cite{griewank2008evaluating} which is a discrete affine
time-variant dynamical system that measures how the parameters of the model would change for infinitesimal variations of the learning rate schedule, after $t$ iterations of the optimization dynamics. 
Notice that the ``translation matrices'' $B_t$ are very sparse, having, at any iteration, only one non-zero column. This means that $[\dot{w}_t]_j$ remains $0$ for all $j\geq t$: $\eta_t$ affects only the future parameters trajectory. Finally, for a single learning rate $\eta_t$, the derivative (a scalar) is 
\begin{align}
\label{eq:truhg}
    \frac{\partial f_T(\eta)}{\partial \eta_t} = [\nabla f_T(\eta)]_{t} =  \left[\left( \prod_{s=t+1}^{T-1} A_s \right) B_{t} \right]^\intercal_t \nabla E(w_T)\\
    \label{eq:truhggd}
    = - \nabla L_t(w_{t})^\intercal \Prange{t+1}{T-1} \nabla E(w_T), 
\end{align}
where the last equality holds true for SGD. Eq. (\ref{eq:truhg}) can be read as the scalar product between the gradients of the training error at the $t$-th step and the objective $E$ at the final iterate, \emph{transformed by} the accumulated (transposed) Jacobians of the optimization dynamics,
shorthanded by $\Prange{t+1}{T-1}$ in (\ref{eq:truhggd}). 
As it is apparent from Eq. (\ref{eq:truhg}), given $w_{t}$, the hypergradient of $\eta_t$ is affected only by the future trajectory 
and does not depend explicitly on $\eta_t$. 

In its original form, where each learning rate is left free to take any permitted value, Problem (\ref{eq:main}) presents a highly nonlinear setup. 
Although in principle it could be tackled by a projected gradient descent method, in practice this is not feasible even for relatively small problems:
evaluating the gradient with forward-mode is inefficient in time since it requires maintaining a large matrix tangent system. Evaluating it with reverse-mode is inefficient in memory since the entire weight trajectory $(w_i)_{i=0}^T$ should be stored\footnote{Techniques based on implicit differentiation~\cite{pedregosa2016hyperparameter,agarwal2017second} 
or fixed-point equations~ \cite{griewank2002reduced} (also known as recurrent backpropagation \cite{pineda1988generalization}) cannot be readily applied to compute $\nabla f_T$ since the training loss $L$ does not depend explicitly on $\eta$.}. Furthermore, it can be expected that several updates of $\eta$ are necessary to reach convergence -- each update requiring the computation of $f_T$ and the entire parameter trajectory in the weight space. Since this approach is computationally very expensive, we turn out attention to online updates where $\eta_t$ is required to be updated online based only on trajectory information up to time $t$. 

\section {Online Gradient-Based Adaptive Schedules}
\label{sec:mu-RTHO}
\label{sec:HD_RTHO_modes}

Before developing and motivating our proposed technique, we discuss two previous methods to compute the learning rate schedule online. 
The real-time hyperparameter optimization (RTHO) algorithm suggested
in~\cite{franceschi2017forward}, reminiscent of stochastic meta-descent \cite{schraudolph1999local},
is based on forward-mode differentiation and uses information from the entire weight trajectory by accumulating partial hypergradients. 
Hypergradient descent (HD), proposed in \cite{baydin_2017_online} and closely related to the earlier work by Almeida \emph{et al.}~\shortcite{almeida1999parameter}, aims at minimizing the loss with respect to the learning rate after one step of optimization. It uses information only from the past and current iterate. 

Both methods implement an update rules of the type
\begin{equation}
\eta_{t} = \max \left[\eta_{t-1} - \beta \Delta\eta_{t}, 0 \right], 
\end{equation} 
where $\Delta \eta_t$ is an online estimate of the hypergradient, $\beta>0$ is a step-size or \emph{hyper-learning rate} and the $\max$ ensures positivity\footnote{Updates could be also considered in the logarithmic space, e.g. by Schraudolph~\shortcite{schraudolph1999local}; 
we find it useful, to let $\eta$ reach 0 whenever needed, offering a natural way to implement early stopping.}. To ease the discussion, 
we omit the stochastic (mini-batch) evaluation of the training error $L$ and possibly of the objective $E$. 

The update rules\footnote{
In \cite{franceschi2017forward}, the hyperparameter is updated every $K$ iterations. 
Here we focus on the case $K=1$ which better allows for a unifying treatment. HD is developed using as objective the training loss $L$ rather than the validation loss $E$. 
We consider here without loss of generality the case of optimizing $E$.
} 
are given by
\begin{align} \label{eq:rtho_hd_deltas}
\Delta^{\mathrm{RTHO}} \eta_t &= \left[ \sum_{i=0}^{t-1} \Prange{i+1}{t-1} B_i \right]^\intercal \nabla E(w_t);
\\ 
\Delta^{\mathrm{HD}} \eta_t &= B_{t-1}^\intercal \nabla E(w_t)
\end{align}
for RTHO and HD respectively, where $\Prange{t}{t-1}\coloneqq I$. Thus $\Delta^{\mathrm{RTHO}} = \Delta^{\mathrm{HD}} + r((w_i, \eta_i)_{i=0}^{t-2})$: the correction term $r$ can be interpreted as an ``on-trajectory approximations'' of longer horizon objectives as we will discuss in Section \ref{sec:proposal}. 

\begin{figure*}
\centering
\includegraphics[width=0.96\textwidth]{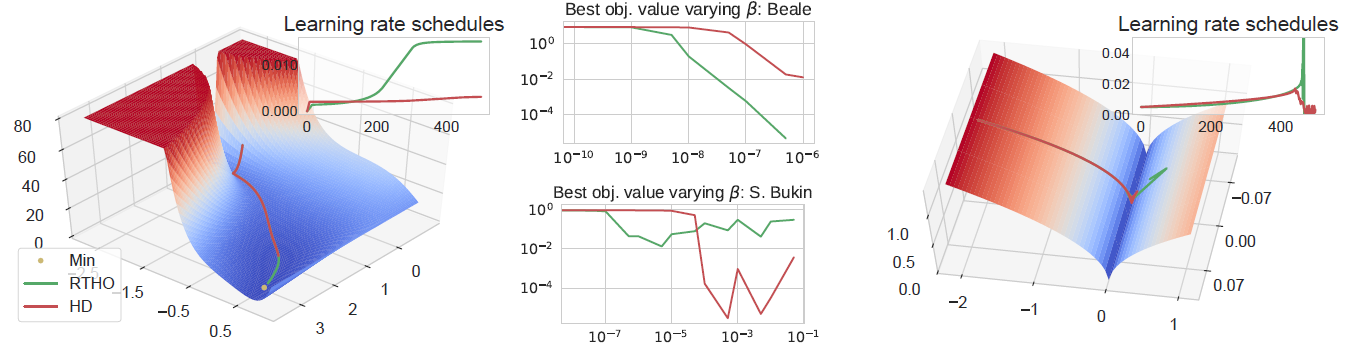}
\vspace{-2mm}
\caption{Loss surface and trajectories for 500 steps of gradient descent with HD and RTHO for Beale function (left) and (smoothed and simplified) Bukin N.6 (right). Center: best objective value reached within 500 iterations for various values of $\beta$ that do not lead to divergence.}
    \label{fig:fm_imgs}
    \label{fig:test_func}
\vspace{-5mm}
\end{figure*}

Although successful in some learning scenarios, we argue that both these update
rules suffer from (different) pathological behaviors, as HD may be
``shortsighted'', being prone to underestimate the learning rate (as noted by Wu \emph{et al.}~\shortcite{wu2018understanding}), while RTHO may be too slow to adapt to sudden
changes of the loss surface or, worse, it may be unstable, with updates growing uncontrollably in magnitude.
We exemplify these behaviors in Figure \ref{fig:fm_imgs}, using two bidimensional test functions\footnote{We use the Beale function defined as $L(x,y) = (1.5 - x + xy)^2 + (2.25-x+xy^2)^2 + (2.625-x+xy^3)^2$ and a simplified smoothed version of Buking N.6: $L(x, y) = \sqrt{  ((y - 0.01 x)^2 + \varepsilon)^{1/2} + \varepsilon}$, with $\varepsilon>0$.} from the optimization literature, where we set $E=L$ and we perform 500 steps of gradient descent.
The Beale function, on the left, presents sharp peaks and large plateaus. RTHO consistently outperforms HD for all probed values of $\beta$ that do not lead to divergence (Figure \ref{fig:test_func} upper center). 
This can be easily explained by the fact that in flat regions gradients are small in magnitude, leading to $\Delta^{\mathrm{HD}} \eta_t$ to be small as well. RTHO, on the other hand, by accumulating all available partial hypergradients and exploiting second order information, is capable of making faster progress. We use a simplified and smoothed version of the Bukin function N.6 to show the opposite scenario (Figure \ref{fig:test_func} lower center and right). Once the optimization trajectory closes the valley of minimizers $y = 0.01 x$, RTHO fails to discount outdated information, bringing the learning rate first to grow exponentially,
and then to suddenly vanish to 0, as the gradient changes direction. HD, on the other hand, correctly damps $\eta$ and is able to maintain the trajectory close to the valley.

These considerations suggest that neither $\Delta^{\mathrm{RTHO}}$ nor
$\Delta^{\mathrm{HD}}$ provide \emph{globally useful} update directions, as large plateaus and sudden changes on the loss surface are common features of  the optimization landscape of neural networks \cite{bengio1994learning,glorot2010understanding}. 
Our proposed algorithm smoothly interpolates between these two methods, as we will present next.

\section{\nostro}  \label{sec:proposal}

In this section, we develop and motivate \nostro, an algorithm for computing LR schedules online during a single training run. This method maintains a moving-average over approximations of~(\ref{eq:truhg}) of increasingly longer horizon, using the past trajectory and gradients to retain a low computational overhead.
Further, we show that RTHO \cite{franceschi2017forward} and HD \cite{baydin_2017_online} outlined above, 
can be interpreted as special cases of \nostro, shedding further light on their behaviour and shortcomings. 

\paragraph{Shorter horizon auxiliary objectives.} For $K>0$, define $g_K(u, \xi)$, with $\xi \in \mathbb{R}_+^{K}$ as
\begin{align}
  \label{eq:class}
  g_K(u, \xi) =  E(u_K(\xi))  \quad \text{s.t.} \quad u_{0} = u, \\
  u_{i+1} = \Phi(u_{i}, \xi_i) \;\; \text{for} \;\; i=[K].
\end{align} 
The $g_K$s  define a class of shorter horizon objective functions, indexed by $K$, 
which correspond to the evaluation of $E$ after $K$ steps of optimization, starting from $u\in\mathbb{R}^d$ and using $\xi$ as the LR schedule\footnote{
Formally, $\xi$ and $u$ are different from $\eta$ and $w$ from the previous sections; later, however, we will evaluate the $g_K$'s on subsequences of the optimization trajectory.}.
Now, the derivative of $g_K$ with respect to $\xi_0$, denoted
$g'_K$, is given by
\begin{align} \label{eq:hg:short}
    g'_K(u, \xi) =  \frac{\partial g_K(u, \xi)}{\partial \xi_0} = \left[B_{0} 
    \right]_{0}^\intercal  \Prange{1}{K-1}   \nabla E(u_{K}) \\
    = - \nabla L(u)^\intercal \Prange{1}{K-1} \nabla E(u_{K}),
\end{align}
where the last equality holds for SGD dynamics.
Once computed on subsets of the original optimization dynamics $(w_i)_{i=0}^T$, the derivative  
reduces for $K=1$
to $g'_1(w_t, \eta_t) = -\nabla E(
w_{t+1})\nabla L(w_t)^\intercal$ (for SGD dynamics), and for $K=T-t$ to $g'_{T-t}(w_t, (\eta_i)_{i=t}^{T-1})
= \left[ \nabla f(\eta) \right]_{t}$.
Intermediate values of $K$ yield cheaper, shorter horizon approximations of (\ref{eq:truhg}).

\paragraph{Approximating the future trajectory with the past.}  
Explicitly using any of the approximations given by $g'_K(w_t, \eta)$
as $\Delta \eta_t$ is, however, still largely impractical, especially for $K\gg 1$. Indeed, it would be necessary to iterate the map $\Phi$ for $K$ steps in the future, with the resulting  $(w_{t+i})_{i=1}^{K}$ iterations discarded after a single update of the learning rate. 
For $K\in[t]$, we may then consider evaluating $g'_K$ \emph{exactly $K$ steps in the past}, that is evaluating $g'_K(w_{t-K}, (\eta_i)_{i=t-K}^{t-1})$. Selecting $K=1$  is indeed equivalent to $\Delta^{\mathrm{HD}}$, which is computationally inexpensive.  However, when past iterates 
are close to future ones 
(such as in the case of large plateaus), using larger $K$'s would allow  in principle to capture longer horizon dependencies present in the hypergradient structure 
 of Eq. 
(\ref{eq:truhg}).
Unfortunately the  computational efficiency of $K=1$ does not generalize to $K>1$, since setting $\Delta \eta_t = g'_K$  would require maintaining $K$ different tangent systems.   

\paragraph{Discounted accumulation of $g'_k$s.} The definition of the $g_K$s, however, allows one to  highlight the recursive nature of the \emph{accumulation} of $g'_K$. Indeed, by maintaining the vector tangent system,  
\begin{align} \label{eq:zeta}
Z_0 &= \left[B_{0}(u_0, \xi_0)\right]_{0} \\
\label{eq:zsys}
Z_{i+1} &= \mu A_i(u_i, \xi_i) Z_i + [B_i(u_i, \xi_i)]_i\;\; \text{for} \;\; i\geq0,
\end{align}
with $Z_i \in \mathbb{R}^d$, computing the moving average
\begin{equation*}
    S_{K,\mu}(u, \xi) = \sum_{i=0}^{K-1} \mu^{K-1-i} g'_{K-i}(u_i, (\xi_j)_{j=i}^{K-1}) \\ = Z_K^\intercal \nabla E(u_K)
\end{equation*}
from $S_{K-1}$ requires only updating (\ref{eq:zeta}) and recomputing the gradient of $E$. The total cost of this operation is $O(c(\Phi))$ per step  both in time and memory using fast Jacobians vector products \cite{pearlmutter1994fast} where $c(\Phi)$ is the cost of computing the optimization dynamics (typically $c(\Phi)= O(d)$). 
The parameter $\mu\in[0,1]$ allows to control how quickly past history is forgotten. One can notice that $\Delta^{\mathrm{RTHO}} \eta_t = S_{t,1}(w_0, (\eta_j)_{i=0}^{t-1})$, while $\mu=0$ recovers $\Delta^{\mathrm{HD}} \eta_t$. Values of $\mu<1$ help discounting outdated information, while as $\mu$ increases
so does the horizon of the hypergradient approximations.  
The computational 
scheme of Eq. (\ref{eq:zeta}) is quite similar to that of forward-mode algorithmic differentiation for computing $\dot{w}$ (see Section \ref{sec:structure} and Eq. (\ref{eq:wdot:upd1})); however, the ``tangent system'' in Eq. (\ref{eq:zeta}), exploiting the sparsity of the matrices $B_t$, only keeps track of the variations with respect to the first component $\xi_0$, drastically reducing the running time. 

Algorithm \ref{algo:nostro} presents the pseudocode of \nostro{}. The runtime and memory requirements of the algorithm are dominated by the computation of the variables $Z$. Being these structurally identical to the tangent propagation of forward mode algorithmic differentiation, we conclude that the runtime complexity is only a multiplicative factor higher than that of the underlying optimization dynamics $\Phi$ and requires two times the memory (see Griewank and Walther \shortcite{griewank2008evaluating}, Sec. 4). 
\begin{algorithm}[tb]
\caption{\small {\bf \nostro{}}; requires $\beta$, $\mu$, $\eta_0[=0]$}
\label{algo:nostro}
\begin{algorithmic}
    \STATE Initialization of $w$ and $Z_0 \gets 0$
    \FOR{$t=0$ {\bfseries to} $T$}
    \STATE $\eta_{t} \gets \max \left[\eta_{t-1} - \beta \Delta\eta_{t}, 0 \right]$ \hfill \COMMENT{Update LR if $t>0$}
	\STATE $Z_{t+1} \gets \mu A_t(w_t, \eta_t) Z_t + [B_t(w_t, \eta_t)]_t$ \hfill \COMMENT{Eq. (\ref{eq:zsys})}
    \STATE $w_{t+1} \gets \Phi_t(w_t, \eta_t)$  \hfill \COMMENT{Parameter update}
    \ENDFOR 
\end{algorithmic}
\end{algorithm}
\section{Optimized Schedules and Quality of \nostro{} Approximations}

\begin{figure*}
\centering
\includegraphics[width=1\textwidth]{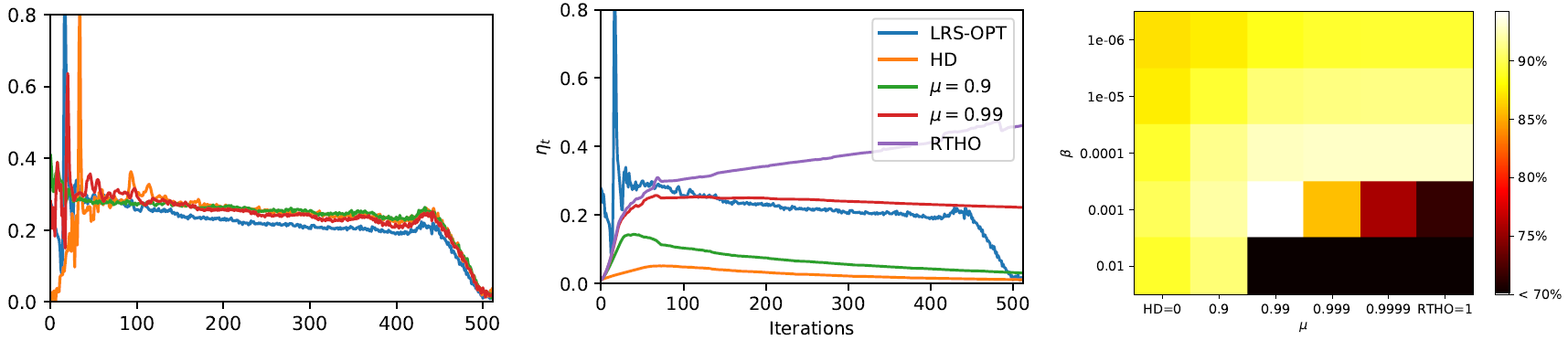}
\vspace{-3mm}
\caption{\small  
    Left: schedules found by LRS-OPT (after 500 iterations of SGD) on 4 different random seeds. Center: comparison between optimized and \nostro{} schedules for one seed, for indicative values of $\mu$. We report the schedule generated with the hyper-learning rate $\beta$ that achieves the best final validation accuracy. Right: Average validation accuracy of \nostro{} over 20 random seeds, for various values of $\beta$ and $\mu$. The best performance of $94.2\%$ is obtained with $\mu=0.99$. For reference, the average validation accuracy of the network trained with $\eta = 0.01\cdot\mathbf{1}_{512}$ is $87.5\%$, while LRS-OPT obtains an average accuracy of $96.1\%$. 
   For $\mu\in[0.9, 1)$, when \nostro{} converges it consistently outperforms HD and it performs at least as well as RTHO, but converges for a wider range of $\beta$.}
    \label{fig:LR_mrtho_opt} \label{fig:qual}
    \vspace{-2mm} 
\end{figure*}

\label{sec:true-vs-approx}
In this section, we empirically compare the optimized LR schedules found by approximately solving Problem~\ref{eq:main} by gradient descent (denoted LRS-OPT), where the hypergradient is given by Eq. (\ref{eq:truhg}), against those generated by \nostro{}, for a wide range of hyper-momentum factors $\mu$ (including HD and  RTHO) and hyper-learning rates $\beta$. We are interested in understanding and visualizing the qualitative similarities among the schedules, as well as the effect of $\mu$ and $\beta$ on the final performance measure. 
To this end, we trained three-layers feed forward neural networks with 500 hidden units per layer on a subset of 7000
MNIST~\cite{lecun1998gradient} images. We used a cross-entropy
loss and SGD as optimization dynamics $\Phi$, with a mini-batch size of 100. 
We further sampled 700 images to form the validation set and defined $E$ to be
the validation loss after $T=512$ optimization steps (about 7 epochs).
For LRS-OPT, we randomly generated different mini-batches at each iteration, to prevent the
schedule from unnaturally adapting to a specific progression of
mini-batches\footnote{We retained, however, the random initialization of the network weights,
to account for the impact that this may have on the initial part of the trajectory. This offers a fairer comparison between
LRS-OPT and online methods, which compute the trajectory only once.}.
We initialized $\eta = 0.01\cdot\mathbf{1}_{512}$ for LRS-OPT and set $\eta_0=0.01$ for \nostro{}, and repeated the experiments for 4 random seeds.

Figure~\ref{fig:qual} (left) shows the LRS-OPT schedules found after 5000 
iterations of gradient descent: 
the plot reveals a strong initialization (random seed) specific behavior of $\eta^*$ for approximately the first 
100 steps. The LR schedule then stabilizes or slowly decreases up until around 50 iterations before the final time, at which point it quickly decreases (recall that, in this setting, all $\eta_i$, including $\eta_0$, are optimized ``independently'' and may take any permitted value). 
Figure \ref{fig:qual} (center) present a qualitative comparison between the offline LRS-OPT schedule and the online ones, for indicative values of $\mu$. Too small values of $\mu$ result in an underestimation of the learning rates, with generated schedules that quickly decay to very small values -- this is in line with what observed in \cite{wu2018understanding}. For too high values of $\mu$ ($\mu=1$ i.e. RTHO \cite{franceschi2017forward} in the figure) the schedules linger or fail to decrease, possibly causing instability and divergence. For certain values of $\mu$, the schedules computed by \nostro{} seems to capture the general behaviour of the optimized ones. 
Finally, Figure \ref{fig:qual} (right) shows the average validation accuracy (rather than loss, for easier 
interpretation) of \nostro{} methods varying $\beta$ and $\mu$. Higher values of $\mu$ translate to higher final 
performances -- with a clear jump occurring between $\mu=0.9$ and $\mu=0.99$ -- but may require a smaller hyper learning rate to prevent divergence.

\section{Experiments} \label{sec:exp}
\begin{figure*}[t]
\centering
\begin{minipage}{0.03\textwidth}
\end{minipage}
\begin{minipage}{0.35\textwidth}
       \includegraphics[width=1.\textwidth, trim={0.5cm 0.0cm 1.5cm 0.5cm}, clip]{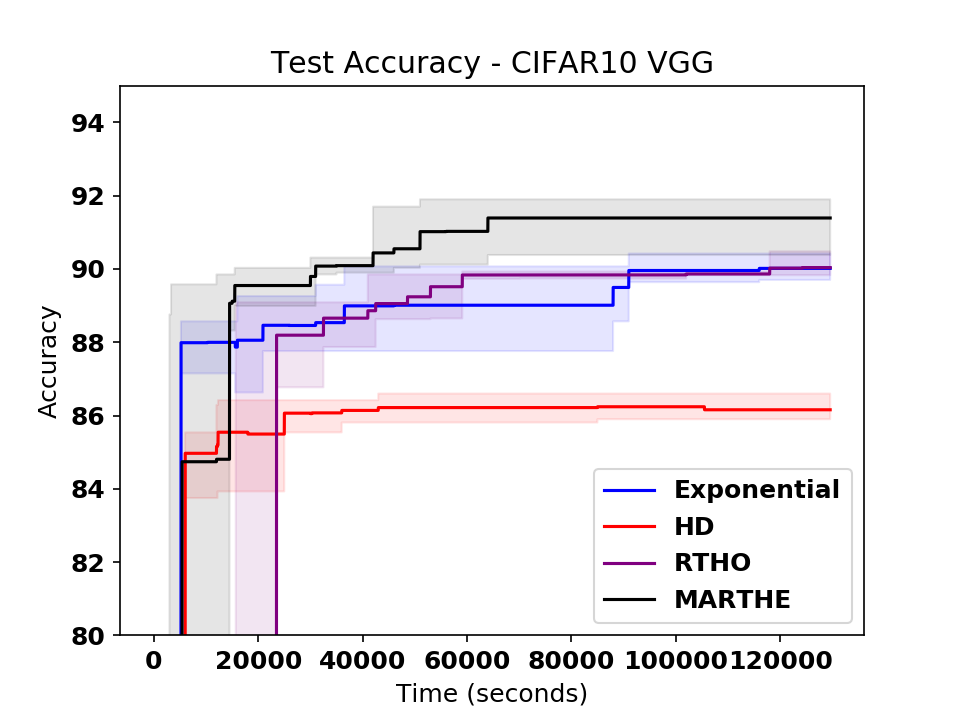}
\end{minipage}
\begin{minipage}{0.03\textwidth}
\end{minipage}
\begin{minipage}{0.35\textwidth}
        \includegraphics[width=1.\textwidth, trim={0.5cm 0.0cm 1.5cm 0.5cm}, clip]{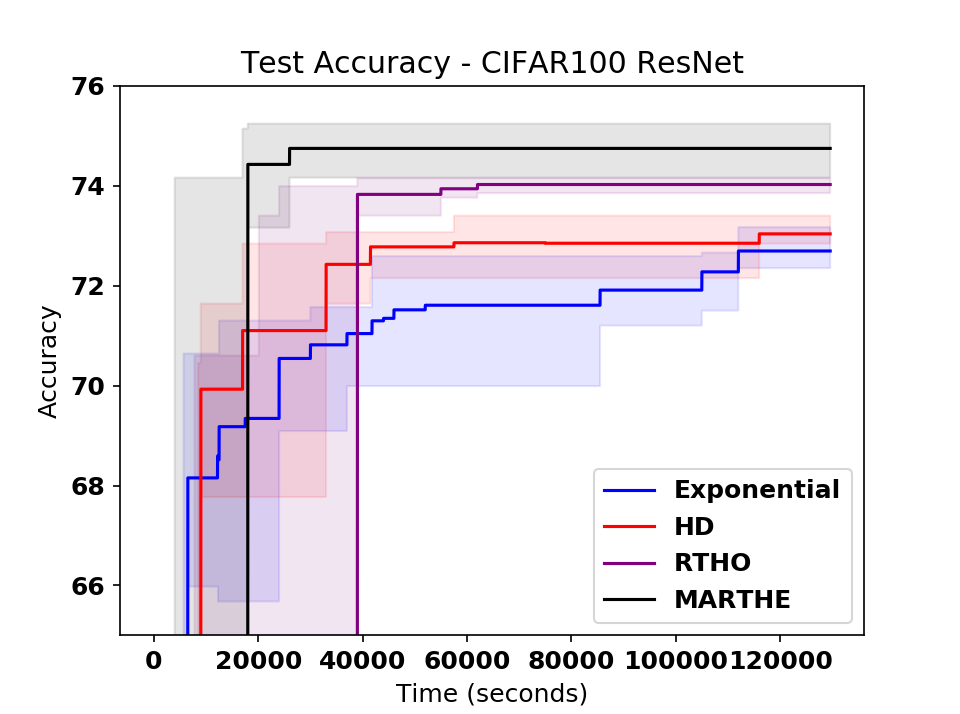}
        
\end{minipage}
\begin{minipage}{0.03\textwidth}
\end{minipage}
\begin{minipage}{0.24\textwidth}
\centering
    \includegraphics[width=0.95\textwidth, trim={0.0cm 0.0cm 0.0cm 0.0cm}, clip]{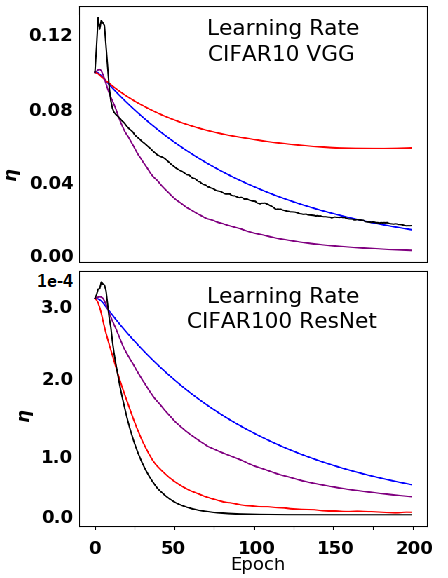}
\end{minipage}
\begin{minipage}{0.03\textwidth}
\end{minipage}
\vspace{-3mm}
\caption{\small  
    Left and center: we randomly draw parameters from each algorithm's configuration space (hyper-hyperparameters) and run the resulting experiments using early stopping with a patience window of 10 epochs. We keep track of the best model (i.e. the model with the highest validation accuracy) found so far and we report the relative test accuracy as a function of time. The solid line represents average accuracy, while shaded regions depict minimum and maximum accuracy across different seeds.
    On the left we show results for the experiments with VGG networks trained on CIFAR10 with SGDM as inner optimization method. The center plot reports experiments with ResNet models trained on CIFAR100 with Adam as optimization dynamics. Right: samples of learning rate schedules that lead to the best found model for each scheduling method and for the relative seed. Experiments on CIFAR10 (top) and CIFAR100 (bottom).}
    \label{fig:cifar_all}
    \vspace{-3mm} 
\end{figure*}

We performed an extensive set of experiments in order to compare \nostro{}, RTHO, and HD. We also considered a classic LR scheduling baseline in the form of exponential decay (Exponential) where the LR schedule is defined by $\eta_t=\eta_0\gamma^t$.
The purpose of these experiments is 
to perform a thorough comparison of various learning-rate scheduling methods, with a focus on those that are (hyper-)gradient based, in the fairest possible manner: 
indeed, these methods have very different running-time per iteration -- HD and Exponential being much faster than \nostro{} and RTHO -- as well as different configuration spaces. It would be unfair to compare them using the number of iterations as computational budget. 
We therefore designed an experimental setup that allowed us to account for it: we implemented a random search strategy over the respective algorithms' configuration spaces and early-stopped each run with a 10-epochs patience window. We repeatedly drew configurations parameters (hyper-hyperparameters) and run respective experiments until a fixed time budget of $36$ hours was reached.
The proposed experimental setting tries to mimic how machine learning practitioners may approach the parameter-tuning problem.

We used two alternative optimization dynamics: SGDM with the momentum hyperparameter fixed to $0.9$ and Adam with the commonly suggested default values $\beta_1 =0.9$ and $\beta_2=0.999$. We fixed the batch size to $128$, the initial learning rate $\eta_0 = 0.1$ for SGDM and $0.003$ for Adam, and the weight decay (i.e. $2$-norm) to $5 \cdot 10^{-4}$. For the adaptive methods, we sampled $\beta$ in $[10^{-3}, 10^{-6}]$ log-uniformly, and for our method, we sampled $\mu$ between $0.9$ and $0.999$. Finally, we picked the decay factor $\gamma$ for Exponential log-uniformly in $[0.9, 1.0]$.

In our first set of experiments, we used VGG-11~\cite{simonyan2014very} 
with batch normalization~\cite{ioffe2015batch} after the convolutional layers, on
CIFAR10~\cite{krizhevsky2014cifar}, and SGDM as the inner optimizer. In the second set of experiments, we used ResNet-18~\cite{He_2016_CVPR} on CIFAR100~\cite{krizhevsky2014cifar}, in this case with Adam. 
For both CIFAR10 and CIFAR100, we used 45000 images as training images and 5000 images as the validation dataset. An additional set of 10000 test images was finally used to estimate generalization accuracy. We used standard data augmentation, including mean-std normalization, random crops and horizontal flips. Gradients were clipped to an absolute value of 100.0.

We kept track of the model with the best validation accuracy found so far, reporting in Figure \ref{fig:cifar_all} (left and center) the relative mean test accuracy (solid line) and minimum and maximum (limits of the shaded regions) across $5$ repetitions. 
Inspecting the figure, it is possible to identify which method is the best performing one for any give time budget, both in average and in the worst/best case scenario. Figures~\ref{fig:cifar_all} (right)
shows examples of the LR schedules obtained by using the different methods.

We performed all experiments using AWS P3.2XL instances, each providing one NVIDIA Tesla V100 GPU. Finally, our PyTorch implementation of the methods and the experimental framework to reproduce the results is available at \url{https://github.com/awslabs/adatune}.
\subsection{Discussion}
In the analysis of our results, we will mainly focus on the
accuracy on the test dataset achieved by different
methods within a fixed time budget. For all the experiments, results
summarized in Figure \ref{fig:cifar_all} show that both Exponential and HD were able
to obtain a reasonably good accuracy within the first 4 hours, while RTHO
and \nostro{} required 6 hours at least to reach the same
level of accuracy. This is due to the fact that the wall-clock time
required to process a single minibatch is different: 
\nostro{} takes approximately 4 times the wall-clock 
time of HD; there is negligible wall-clock time difference 
between \nostro{} and RTHO or between HD and Exponential.
\nostro{} was able to surpass all the other
methods consistently after 10 hours.

Our experimental protocol resulted in HD and Exponential getting
more trials compared to RTHO and \nostro{} (in average around 24 trials for the first two compared to 8 of RTHO and \nostro{}).  Despite the fact
that \nostro{} could only afford fewer trials, it could still achieve better performance,
suggesting that it is able to produce better learning rate
schedules more reliably. Moreover, \nostro{} maintains a better peak accuracy compared
to RTHO showing the effectiveness of down-weighting outdated information.

Our experimental setup helped us investigate the robustness of
the methods with respect to the choice of the
hyper-hyperparameters\footnote{We note that it is not common in the
existing literature to mention the necessity of tuning
hyper-hyperparameters for adaptive learning-rate methods, although different datasets and/or networks may require some tuning that may strongly affect the results.}. To that end, we can see from Figure~\ref{fig:cifar_all} (left) that the average and worst case test set accuracies (measured across multiple seeds) of \nostro{} are better in comparison to the other methods. This is a strong indication that \nostro{} demonstrated superior adaptability with respect to different hyper-hyperparameters and seeds compared to other methods. This is also reflected by the result that \nostro{} outperforms other strategies on a consistent basis if given a sufficient time budget (4-6 hours in our experiments): the higher computational cost of \nostro{} is outbalanced by the fact that it needs fewer trials to reach the same performance level of faster methods like Exponential or HD.

Overall, our experiments reveal that RTHO and \nostro{} provide better performance, giving a clear indication of the importance of the past information. Due to its lower computational overhead, Exponential should be still preferred under tight budget constraints, while \nostro{} with $\mu\in[0.9, 0.999]$ should be preferred if enough time is available. 

\section{Conclusion}
\label{sec:conclusions}
Finding a good learning rate schedule is an old but crucially important issue in machine learning.
This paper makes a step forward, analyzing previously proposed online gradient-based methods and introducing a more general technique to obtain performing LR schedules based on an increasingly long moving average over hypergradient approximations. \nostro{} interpolates between HD and RTHO, and its implementation is fairly simple within modern automatic differentiation, adding only a moderate computational overhead over the underlying optimizer complexity.

In this work, we studied the case of optimizing the learning rate schedules for image classification tasks; we note, however, that \nostro{} is a general technique for finding online hyperparameter schedules (albeit it scales linearly with the number of hyperparameters), possibly implementing a competitive alternative in other application scenarios, such as tuning regularization parameters \cite{luketina2016scalable}. We plan to further validate the method both in other learning domains for adapting the LR and also to automatically tune other crucial hyperparameters. 
We believe that another interesting future research direction could be to derive or learn adaptive rules for $\mu$ and $\beta$.

\newpage
\cleardoublepage
\bibliographystyle{named}
\bibliography{lrpp_biblio}

\end{document}

%% file: math_commands.tex
\usepackage{amsmath,amsfonts,bm}

\def\eqref#1{equation~\ref{#1}}

\def\1{\bm{1}}

\DeclareMathAlphabet{\mathsfit}{\encodingdefault}{\sfdefault}{m}{sl}
\SetMathAlphabet{\mathsfit}{bold}{\encodingdefault}{\sfdefault}{bx}{n}

